\documentclass{article} 
\usepackage{iclr2023_conference,times}

\usepackage{hyperref}
\usepackage{url}

\usepackage{algorithm}
\usepackage{algorithmic}
\usepackage[switch]{lineno}

\usepackage{bm}
\usepackage{multirow}
\usepackage[figuresright]{rotating}
\usepackage{bbding}
\usepackage{amsmath}
\usepackage{graphicx}
\usepackage{natbib}

\iclrfinaltrue
\title{Logical Bias Learning for Object Relation Prediction}

\author{Xinyu Zhou\dag, Zihan Ji\dag  \& Anna Zhu\thanks{Corresponding Author, \dag Equal Contribution.}\\
Department of Computer and Artificial Intelligence\\
Wuhan University of Technology\\
\texttt{\{297932, jizihan, annazhu\}@whut.edu.cn} \\
}

\begin{document}

\maketitle

\begin{abstract}
Scene graph generation (SGG) aims to automatically map an image into a semantic structural graph for better scene understanding. It has attracted significant attention for its ability to provide object and relation information, enabling graph reasoning for downstream tasks. However, it faces severe limitations in practice due to the biased data and training method. In this paper, we present a more rational and effective strategy based on causal inference for object relation prediction. To further evaluate the superiority of our strategy, we propose an object enhancement module to conduct ablation studies. Experimental results on the Visual Gnome 150 (VG-150) dataset demonstrate the effectiveness of our proposed method. These contributions can provide great potential for foundation models for decision-making.

\end{abstract}

\section{Introduction}

\label{Intro}

Scene graph generation (SGG) aims to generate a comprehensive textual graph that includes nodes representing object classes and edges denoting their pairwise relations. It has attracted significant attention due to its support for graph reasoning. Besides, it is also a good method to automatically generate pre-training data for foundation models. However, in recent years, there has been an evident decline in the number of cross-modal methods based on scene graphs. This suggests that the SGG task has deviated from practice, which is a confusing phenomenon as graph structures are widely and increasingly used in various tasks. After conducting a thorough investigation, we determine that the primary cause of this decline is the inefficiency of dealing with the relation bias problem.

\begin{figure*}[!htb]
  \centering
  \includegraphics[width=0.95\linewidth]{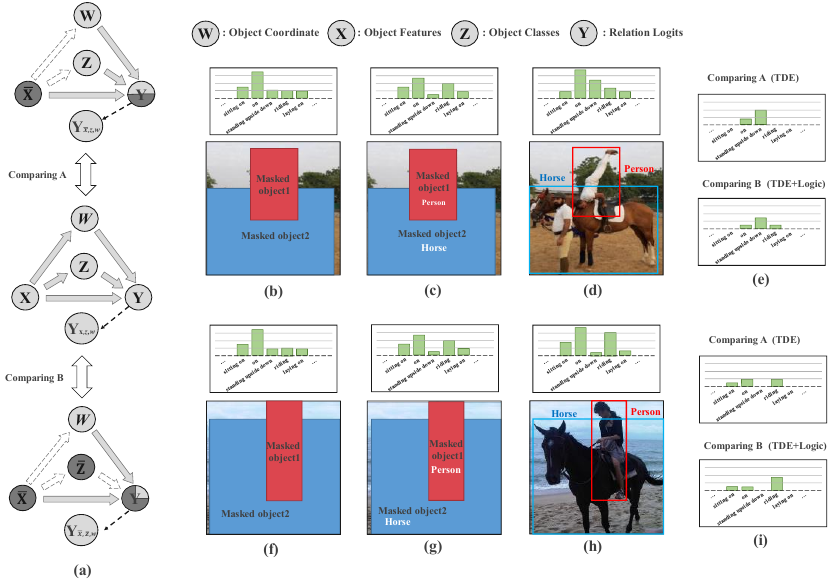}
  \caption{The illustration of the LBL strategy. (a) The causal graph for TDE w/o logic calculation. (b)\&(f) Scenarios with masked object pairs' bounding boxes. (c)\&(g) Scenarios with masked object pairs' bounding boxes and classes. (d)\&(h) Scenarios with masked object pairs' bounding boxes, classes and visual features. (e)\&(i) are the TDE w/o logic scores for relation prediction of the two objects in (d)\&(h), respectively.}
  \label{fig2}
\end{figure*}
\vspace{-0.2cm}

The biased predictions arise from the long-tail distribution of data and the inclusion relationship among relations. In other words, this problem comes from a statistical perspective. Unfortunately, most existing methods manage to solve it via complex model designs, which are too specific and inefficient to be used in practice. Therefore, our proposal is to find a simple yet effective method. Subsequently, a superb strategy based on causal inference \cite{glymour2016causal} is proposed, which is motivated by TDE \cite{tang2020unbiased} and the phenomenon that \textbf{students would ask their teacher for help if they are confused.}

As shown in Fig. \ref{fig2}, given some choices, e.g., ``on", ``riding", ``standing upside down", you are required to describe the relation between two objects in the image. For Fig. \ref{fig2} (b) and (f), most people would prefer to choose ``on'' between the two masked objects without visual and class information. It is inferred only from the object layout, namely the coordinate. Although relations like ``on'' and ``near'' are correct, they are useless for reasoning since the information is too rough. Naturally, we call this kind of bias ``bad bias".

If providing the object class information about the masked regions, i.e., given the classes for object1 and object2 as ``person'' and ``horse'' respectively as in Fig. \ref{fig2} (c) and (g), we have the alternative word ``riding'' to represent their relation in this counterfactual scene. Because it matches our intuition that ``riding'' is common in the combination of ``person'' and ``horse''. This inference comes from our common sense, which is rational \cite{simon1990bounded} and in line with most cases. Hence, it is ``good bias" relating to our logic thinking. We call it ``logic bias". It could provide extra knowledge and help with judgement when facing many uncertain choices.

Additionally providing the visual information, as shown in Fig. \ref{fig2}(h), the conclusion may be the same as the former. However, there also exists the case in Fig. \ref{fig2} (d), where the scenario is unusual and the relation is described as ``person standing upside down horse''. Total Direct Effect (TDE) \cite{tang2020unbiased} is a strategy to solve this issue by empowering machines with the ability of counterfactual causality \cite{pearl2018book}.

Fig. \ref{fig2} (a) presents the underlying causal graphs of the above three alternate scenes. The arrow in $x$ → $y$ indicates that node $y$ is caused by node $x$. In relation prediction, there are three factors: object visual features ($X$), classes ($Z$) and coordinates ($W$); the faded links in the upper and bottom graphs denote the wiped-out factors are no longer caused by or affect their linked factors. These graphs offer an algorithmic formulation to calculate TDE.
 
The original TDE-based method \cite{tang2020unbiased} predicts two scores. One is the relation prediction considering object visual, class, and coordinate information (e.g., Fig. \ref{fig2} (d) and (h), represented by $X$+$Y$+$Z$). The other only considers object class and coordinate information (e.g., Fig. \ref{fig2} (c) and (g), represented by $Y$+$Z$). The final score is their subtraction (Comparing A), aiming to predict the relations only through visual information ($X$) of subject and object without extra prior context. This operation can effectively reduce the ``bad bias" and have a good effect for scenarios like Fig. \ref{fig2} (d). However, it \textbf{simultaneously reduce} the ``logical bias" for the common cases like Fig. \ref{fig2} (h). Therefore, TDE may generate uncertainty since it reduces both bad bias and logic bias. The uncertainly predicted score are flatten as shown in Fig. \ref{fig2} (i) Comparing A.


To address this issue, we propose a novel prediction strategy that utilizes knowledge when we encounter uncertainty estimation by TDE. We refer to this strategy as logical bias learning (LBL): when the results from pure visual information are uncertain for decision-making, try to use prior knowledge. It imitates the real scenario as aforementioned: students (TDE: $X$, Comparing A in Fig. \ref{fig2}(a))  would ask their teacher (TDE + Logic: $X$ + $Z$, Comparing B in Fig. \ref{fig2}(a)) for help if they are confused (uncertain). Hence, as shown in Fig. \ref{fig2} (i), using TDE plus logic knowledge strategy when facing uncertainty predictions via the original TDE method, we could get results suppressing bad bias while highlighting the real one that matches our common sense. It perfectly corresponds to the logical reasoning process of humans.

Moreover, we are curious about the potential of this method, i.e., is it possible for normal students (bad performance on $X$) to surpass intelligent students (high performance on $X$) after learning this reasoning method (LBL)? To explore this, we furtuher present an agnostic object feature enhancement module (OEM). Current mainstream methods detect objects with bounding boxes, which may contain redundant and incorrect information from the background and other objects. An instance demonstrating this issue can be observed in Fig. \ref{fig2} (d), where the horse's bbox includes a partial person. This would seriously affect both detection and relation prediction. Inspired by the fact that text representation is much purer compared to images, OEM considers the object class as a query and enhances the targeted visual representations within bboxes through cross-attention. Meanwhile, deformable convolution \cite{dai2017deformable} is employed to effectively extract features from irregular objects. Afterwards, the feature patches are further processed through fine-grained attention, depending on their weights in the attention map.

In summary, our contributions are summarized as follows: \textbf{1)} To solve the relation bias problem efficiently, a novel and effective prediction strategy, LBL, is proposed, which is deeply in line with human logical reasoning. Moreover, we present an object enhancement module to further verify the effectiveness of this strategy, demonstrating ``normal students" can also outperform ``intelligent students". Note that LBL has potential for use in any model for decision-making. \textbf{2)} Experiments on VG-150 indicate we make considerable improvements over the previous state-of-the-art methods. 

\section{Related Work}
\noindent
\textbf{Scene Graph Generation.} SGG aims to generate comprehensive summary graphs for images. It was first proposed by \cite{johnson2015image} in the cross-modal retrieval task. The increasing attention attracted by SGG shows its potential to support the image reasoning tasks \cite{zhou2023scene, yang2019auto, liang2021graphvqa, zhou2022disentangled, nguyen2021defense}. There are two stages for the development process of the SGG. Early methods mainly focused on better visual networks \cite{yin2018zoom, tang2019learning}. After the bias problem was proposed by \cite{zellers2018neural}, many researchers turned to struggle for it \cite{tang2020unbiased, yang2022panoptic}. However, the large cost of existing methods makes this problem not well solved and still far from practice. 

\noindent
\textbf{Unbiased Training.} There are two mainstream methods to solve the bias problem. 1) Labeling a new dataset or resampling existing ones. Like \cite{geirhos2018imagenet}, it posits that the primary cause of bias in SGG lies in the training data. This viewpoint is valid, but the high cost of annotation cannot be ignored. 2) Fusing the prior distribution of training sets. This category, such as \cite{li2022ppdl, lin2022ru}, considers that incorporating the relation distribution from the training set into testing can help eliminate bias. But, the inadequacies of this approach become increasingly pronounced as the dataset undergoes changes. Besides them, there are few methods \cite{tang2020unbiased, dong2022stacked} that take a different approach and are effective in addressing this problem. Our proposed approach falls within this category as well.

\section{Method}

\begin{figure}
  \centering
  \includegraphics[width=\linewidth]{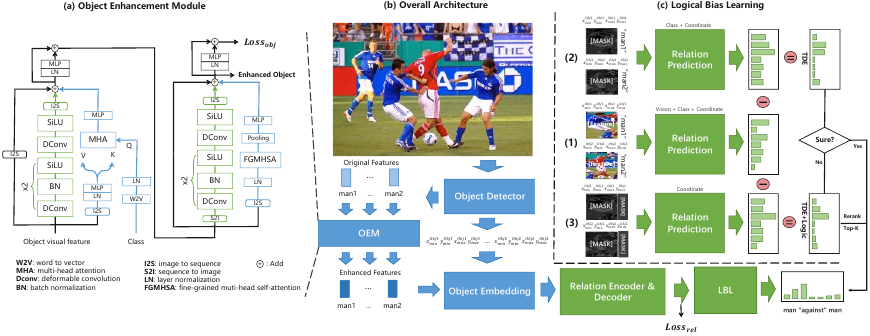}
  \caption{(a) The proposed object enhancement module. (b) The overall architecture of our model. (c) An illustration of the LBL strategy.}
  \label{fig5}
\end{figure}

In this section, we introduce the proposed method in detail. The overall pipeline is shown in Fig. \ref{fig5}. Given an image as input, we extract objects through the object detector. The detected object features are enhanced by the proposed OEM module, which consists of MLP, deformable convolution \cite{dai2017deformable} and multi-head self-attention (MHSA)\cite{vaswani2017attention} blocks. Our model projects representations of entities (in our case, enhanced objects) as vectors in a learned common embedding space. Then, we adopt MOTIFS \cite{zellers2018neural} as the encoder and a fully connected layer as the decoder to the list of projected features for relation probability prediction. The LBL strategy is applied to the estimated relation scores for final relation verification.

\subsection{Object Enhancement Module}\label{OEM}
This module is intended to extract the important semantic information in the detected bounding box and achieve the similar effect of instance segmentation. We fuse the linguistic modal information of object class to refine the bounding box-level visual features through cross-attention operations.

For the first layer, we embed the object class word $x^{word}_{obj}$ (ground-truth when training and predicted one during test) by a 300-d FastText vector $x^{ft}_{obj}$. Then, we regard the text information $x^{ft}_{obj}$ as Q(uery), the visual feature $x^{vf}_{obj}$ as K(ey), and V(alue). To match their dimension, (64, 64)-d object visual feature is first divided into 8$\times$8 patches (each patch has 64-d). Then they are layer normalized and projected to 300-d. Finally, we project the (64, 300)-d matrix back to (64, 64)-d then to 4096-d for residual connection. The cross-attention can be summed up as:
\begin{equation}
\label{eq1}
Attention = \mathrm{Softmax} (\frac{x^{ft}_{obj}{x^{vf}_{obj}}^{\mathsf{T}}}{\sqrt{300}})  x^{vf}_{obj}.
\end{equation}

Some works \cite{pan2022integration, xu2021vitae} have shown the effectiveness of the integration of attention and convolution for better feature representation. Therefore, we refer to the convolution part of ViTAE \cite{xu2021vitae} but replace the normal convolution with the deformable convolution \cite{dai2017deformable} for better concentration on the important parts of objects. The input and output dimensions of MLP remain the same (i.e., 4096 dimensions).

The second layer is a repetition of first layer except the attention mechanism. We use fine-grained multi-head self-attention (FGMHSA) to further extract the fine-grained features of object regions. Top-K important patches are selected based on the attention weights of patch tokens, and then they are split into smaller ones (one fourth) for better representation in a finer granularity. Further, these small patches are upsampled back to original size and tokenized as the input for attention operation. The output of FGMHSA contains more than 64 tokens. They are passed through a pooling layer and projected to 4096-d enhanced object feature $x_{eobj}^{vf}$. We use MLP to project the enhanced feature into $p$-d $x_{eobj}^{class}$ for the classification task. The cross-entropy loss is adopted:
\begin{equation}
\label{eq2}
Loss_{obj} = -\sum_{i=1}^{p} y_{i}\log({x_{eobj}^{class}}_{i}).
\end{equation}
\noindent where $y$ is a $p$-d one-hot code denoting the ground truth of object class.

\subsection{Object Embedding}
\noindent
\textbf{Object Embedding}. We project the enhanced object visual feature $x_{eobj}^{vf}$, embedded class feature $x^{ft}_{obj}$, and the 4-d bounding box coordinate $x^{bbox}_{obj}$: [$x_{min}/W$, $y_{min}/H$, $x_{max}/W$, $y_{max}/$ $H$] into the $d$-d space with three learned linear transforms (where $d$ is 2048). $W$ and $H$ are image width and height respectively. They are summed up as the final object embedding ${x_{obj}}$ as:
\begin{equation}
\label{eq3}
x_{obj} = LN(W_{1}x_{eobj}^{vf}+ W_{2}x^{ft}_{obj})+LN(W_{3}x^{bbox}_{obj}),
\end{equation}

\noindent
where $W_{1}$, $W_{2}$ and $W_{3}$ are learned projection matrices. LN(·) is layer normalization \cite{ba2016layer}, added on the output of the linear transforms.

\subsection{Relation Predication}
For a given image, after detection and feature embedding, we could get a set of objects $\{\bm{x}_{obj}^{m}\}_{1:M}$.

\noindent
\textbf{Relation Encoder}. The MOTIFS (BiLSTMs) \cite{zellers2018neural} is used to encode the objects features. The encoded object is expressed as $x_{i} = BiLSTMs(x_{obj}^{i})$. 

\noindent
\textbf{Relation Decoder}. For each object pair, 4096-d union features $u_{ij}$ are extracted from their overlapped rectangle region to better utilize the context for relation prediction. We concatenate the encoded subject $x_{i}$ and object $x_{j}$ as [$x_{i}$; $x_{j}$], and then project this feature into 4096-d space to fit their union scale. Finally, we use a fully connected layer to predict their relation $R_{ij}$:

\begin{equation}
\label{eq6}
R_{ij}= \mathrm{argmax} (\mathrm{Softmax} (W_{8}(W_{7}([x_{i};x_{j}]) \odot u_{ij})))
\end{equation}

\noindent
where $\odot$ indicates the element-wise product. The prediction loss is implemented in cross entropy as \eqref{eq2}.

\subsection{Logical Bias Learning (LBL)}

The unbiased prediction lies in the difference between the observed factual outcome and its counterfactual alternate. The factual aspect contains object visual features and the context, i.e., their belonging classes and position relations. While the counterfactual aspect removes the real visual features. Fig. \ref{fig5} (c) shows the comparison between them. In Fig. \ref{fig5} (c), (1) represents the prediction result $Y_{x,z,w}$ using the vision $(x)$ $+$ class $(z)$ $+$ bbox $(w)$ features of objects; (2) shows the result $Y_{\bar{x},z,w}$ of using class $(z)$ $+$ bbox $(w)$ features; and (3) is the relation prediction distribution $Y_{\bar{x},\bar{z},w}$ of adopting the bbox $(w)$ features only. For the mask operation, we replace the original feature with a dummy value ($\bar{x}$ or $\bar{z}$), which is termed intervention in causal inference \cite{glymour2016causal}. Only the process of (1) is involved in the training period. Following the proposed LBL strategy, we obtain the TDE with logic (teacher): $Y_{T}$ = $Y_{x,z,w} - Y_{\bar{x},\bar{z},w}$ and the TDE without logic (student): $Y_{S}$ =$Y_{x,z,w} - Y_{\bar{x}, z, w}$. When the prediction $Y_{S}$ encounters uncertainty, we get the result from $Y_{T}$ to re-rank. Otherwise, we directly use the result $Y_{S}$. The uncertainty is defined as: the predicted variance $V^{K}_{p}$ of top-K relations is smaller than the averaged $V^{K}$ in the training sets. In other words, if the confidence of the top-K results of $Y_{S}$ is similar, then go for $Y_{T}$ to re-rank the top-K relations. The final unbiased logits of $Y$ is formatted as:

\begin{equation}
\label{eq7}
\begin{cases}
 Y = Y_{S} & ,\text{ if } V^{K}_{p} \ge V^{K} \\ 
 Y = \mathrm{Rerank}(Y^{0:K}_{S} | Y_{T}) + Y^{K:N}_{S} & ,\text{ if } V^{K}_{p} < V^{K}
\end{cases}
\end{equation}

\noindent
where $N$ is the total number of relations. Note that $K$ is set to 3 experimentally.

\section{Experiments}

\subsection{Experimental Settings}
\noindent
\textbf{Datasets.} 
The experiments of SGG are conducted on two datasets, VG-150. We follow \cite{zellers2018neural, tang2020unbiased, dong2022stacked} to sample a 5k validation set from training set of VG-150 for parameter tuning.\\
\noindent
\textbf{Implementation details.} Following \cite{tang2020unbiased, zhang2022fine}, we employ a pre-trained Faster R-CNN \cite{ren2015faster} with ResNeXt-101-FPN \cite{xie2017aggregated} backbone for object detection. The BiLSTMs is used for relation encoding and a single fully connected layer for decoding. The top-K refers to the top-3 for the certain condition. The top-K important patches refer to the top 50\% patches for the OEM. The 3$\times$3 kernel size is adopted for deformable convolution. Our model is implemented on the Pytorch platform with three RTX A5000 GPUs. We adopt the AdamW optimizer, set the batch size to 12, the initial learning rate to 1e-3 with the weight decay of 1e-4, and a linear decrease scheduler for a total of 40k steps.\\
\noindent
\textbf{Evaluation Metrics.}
We use mean Recall@K (\textbf{mR@K}), a widely used evaluation metric which computes the fraction of times the correct relation is predicted in the top K confident relation prediction, as the metrics for the following three tasks: 1) Predicate Classification (\textbf{PredCls}) provides objects with their corresponding bounding boxes, and requires models to predict the relation of the given pairwise objects; 2) Scene Graph Classification (\textbf{SGCls}) provides the ground-truth object bounding boxes, and needs the models to predict their classes and their pairwise relations. 3) Scene Graph Detection (\textbf{SGDet}) asks the models to detect all the objects and their bounding boxes, as well as predict the relationships of pairwise objects.

\begin{table*}
\caption{Performance (\%) of our method and other SOTA methods on VG-150.}
\label{tab3}\centering
\resizebox{\linewidth}{!}{
\begin{tabular}{ c l | c c c | c c c | c c c }
\hline
\multicolumn{2}{c|}{\Large\multirow{2}*{Methods}} &  \multicolumn{3}{c|}{\bfseries PredCls} & \multicolumn{3}{c|}{\bfseries SGCls}  &  \multicolumn{3}{c}{\bfseries SGDet}\\

 {}& {}& {\bfseries mR@20} & {\bfseries mR@50} & {\bfseries mR@100} & {\bfseries mR@20} & {\bfseries mR@50} & {\bfseries mR@100} & {\bfseries mR@20} & {\bfseries mR@50} & {\bfseries mR@100} \\
 
\hline
\multirow{5}*{\rotatebox{90}{Specific}} 
 & IMP \cite{xu2017scene} & - & 9.8 & 10.5 & - & 5.8 & 6.0 & - & 3.8 & 4.8 \\
 & KERN   \cite{chen2019knowledge} & - & 17.7 & 19.2 & - & 9.4 & 10.0  & - & 6.4 & 7.3\\
 & GBNet  \cite{zareian2020bridging} & - & 22.1 & 24.0 & - & 12.7 & 13.4 & - & 7.1 & 8.5 \\ 
 & PCPL \cite{yan2020pcpl} & - & 35.2 & 37.8 & - & 18.6 & 19.6 & - & 9.5 & 11.7 \\ 
 & BGNN \cite{li2021bipartite} & - & 30.4 & 32.9 & - & 14.3 & 16.5 & - & 10.7 & 12.6 \\
\hline
\multirow{12}*{\rotatebox{90}{Model-Agnostic}} 
 & Motif \cite{zellers2018neural}  & 11.7 & 14.8 & 16.1 & 6.7 & 8.3 & 8.8  & 5.0 & 6.8 & 7.9 \\
 & \quad -TDE \cite{tang2020unbiased}  & 18.5 & 25.5 & 29.1 & 9.8 & 13.1 & 14.9 & 5.8 & 8.2 & 9.8\\
 & \quad -IETrans \cite{zhang2022fine} & - & 35.8 & \bfseries39.1 & - & \bfseries21.5 & \bfseries22.8 & - & 15.5 & 18.0\\ 
 & \quad -GCL \cite{dong2022stacked} & \bfseries30.5 & \bfseries36.1 & 38.2 & \bfseries18.0 & 20.8 & 21.8 & 12.9 & 16.8 & 19.3\\
 & \quad -OEM+LBL \bfseries(ours)  & 27.4 & 32.3 & 35.5 & 16.9 & 19.7 & 21.0  & \bfseries13.1 & \bfseries17.1 & \bfseries19.7 \\
 
\cline{2-11}
 & VCTree  \cite{tang2019learning}  & 13.1 & 16.7 & 18.1 & 9.6 & 11.8 & 12.5 & 5.4 & 7.4 & 8.7 \\
 & \quad -TDE \cite{tang2020unbiased} & 18.4 & 25.4 & 28.7 &  8.9 & 12.2 & 14.0 &  6.9 & 9.3 & 11.1 \\
 & \quad -IETrans \cite{zhang2022fine} & - & 37.0 & \bfseries39.7 & - & 19.9 & 21.8 & - & 12.0 & 14.9\\
 & \quad -GCL \cite{dong2022stacked} & 31.4 & \bfseries37.1 & 39.1 & \bfseries19.5 & \bfseries22.5 & 23.5 & 11.9 & 15.2 & 17.5\\
 & \quad -OEM+LBL \bfseries(ours) & \bfseries29.6 & 34.9 & 38.5& 17.6 & 20.8 & \bfseries24.0  & \bfseries13.2 & \bfseries16.7 & \bfseries18.1\\
\cline{2-11}
\hline

\end{tabular}}
\end{table*}

\subsection{Comparing with Other Methods}
Since our proposed modules can be plugged into any other SGG approach, we compare our method with both specific and agnostic SOTA models. As shown in Tab. \ref{tab3}, we report the specific ones: IMP \cite{xu2017scene}, KERN \cite{chen2019knowledge}, GBNet \cite{zareian2020bridging}, PCPL \cite{yan2020pcpl} and BGNN \cite{li2021bipartite}; and agnostic ones based on MOTIFS \cite{zellers2018neural} and VCTree \cite{tang2019learning}: TDE \cite{tang2020unbiased}, IETrans \cite{zhang2022fine} and GCL \cite{dong2022stacked}. On the widely used OCR-free dataset VG-150, we achieve compariable performance on PredCls and SGcls, and we establish a new state-of-the-art on SGDet, which is the most important metric for being applied to practice.

\subsection{Ablation Study}

To verify the effectiveness of our proposed modules, we conducted ablation experiments on models with and without OEM module and LBL strategy. Since LBL strategy contains two parts, i.e., TDE and TDE plus logic, to calculate the final prediction. We also compare the results with only using TDE or TDE+logic. (Note that TDE here is reproduced version)

\begin{table}[htb]
\caption{Ablations for various modules on VG-150}
\label{tab2}\centering
\scalebox{0.8}{
\resizebox{\linewidth}{!}{
\begin{tabular}{ c c c c |c c | c c | c c}
\hline
  \multicolumn{4}{c|}{\bfseries Module} & \multicolumn{2}{c|}{\bfseries PredCls (\%)} & \multicolumn{2}{c|}{\bfseries SGCls (\%)} & \multicolumn{2}{c}{\bfseries SGDet (\%)} \\
   {OEM} & {TDE} & {Logic} & {LBL} & {\bfseries mR@50} & {\bfseries mR@100} & {\bfseries mR@50} & {\bfseries mR@100} & {\bfseries mR@50} & {\bfseries mR@100}\\
\hline
  \XSolidBrush & \Checkmark & \XSolidBrush & \XSolidBrush & 23.6 & 27.7 & 12.4 & 14.1 & 8.0 & 9.6 \\
  \XSolidBrush & \Checkmark & \Checkmark & \XSolidBrush & 20.6 & 26.4 & 11.8 & 15.6 & 10.9 & 12.4 \\
  \Checkmark & \Checkmark & \XSolidBrush & \XSolidBrush & 24.0 & 27.7 & 12.6 & 16.4 & 11.6 &  14.6\\
  \Checkmark & \Checkmark & \Checkmark & \XSolidBrush & 26.3 & 30.4 & 15.3 & 18.2 & 12.6 & 15.0\\ 
  \XSolidBrush & \Checkmark & \Checkmark & \Checkmark & 28.6 & 32.5 & 16.6 & 19.6 & 12.8 & 15.4 \\ 
  \Checkmark & \Checkmark & \Checkmark & \Checkmark & 
  \bfseries32.3 & \bfseries35.5 & \bfseries19.7 & \bfseries21.0 & \bfseries17.1 & \bfseries19.7 \\
 
\hline
\end{tabular}}
}
\end{table}

As shown in Tab. \ref{tab2}, the performance of independent TDE with Logic and TDE (w/o Logic) is similar. However, after implementing our LBL strategy, significant improvements are made among the three tasks. Besides, LBL w/o OEM is better than TDE with OEM shows the powerful ability of the LBL strategy. It also verifies our aforementioned assumption that \textbf{with teachers' help (logic), normal students (TDE) may surpass intelligent students (TDE + OEM).}

\subsection{Qualitative Study}

 \begin{figure}[htb]
  \centering
  \includegraphics[width=0.85\linewidth]{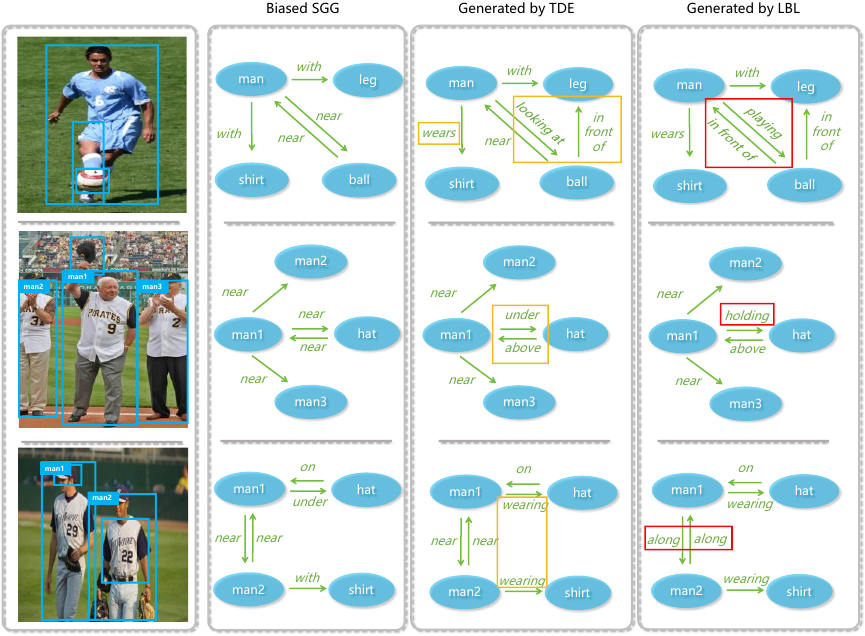}
  \caption{Examples of the visualized normal SGG and the unbiased ones generated by TDE\&LBL.} 
  \label{fig7}
\end{figure}

We present three kinds of results qualitatively in Fig. \ref{fig7}. The blue circles represent objects. Relations are denoted by the green arrows. We can see that both TDE and LBL can well overcome the biased problem in (a): from ``near'' to ``under'' and ``above'' (labeled by yellow boxes). However, in scenarios (b) and (c), the aforementioned uncertain prediction by TDE takes place, but our strategy LBL can address it well (labeled by red boxes).

\subsection{Specific Performance}

\begin{table*}[!htb]
\renewcommand\arraystretch{1.1}
\caption{Specific results of \textbf{mR@100} on VG-150 }\label{tabx}\centering
\resizebox{0.9\textwidth}{!}{
\begin{tabular}{ c | c c c | c | c c c}
\hline
{\Large\multirow{2}{*}{\bfseries Relation Class}} &  \multicolumn{3}{c|}{\bfseries TDE as basic} & {\Large\multirow{2}{*}{\bfseries Relation Class}} & \multicolumn{3}{c}{\bfseries TDE as basic}\\
 {} & {\bfseries +OEM+LBL} & {\bfseries +OEM} & {\bfseries TDE only} & {} & {\bfseries +OEM+LBL} & {\bfseries +OEM} & {\bfseries TDE only}   \\
\hline
above	& 0.1221	& 0.1471	& 0.0821
& across  & 	0.0000	& 0.0000	& 0.0000\\
\bfseries against &	\bfseries 0.1403	& 0.081	    & 0.0000
& \bfseries along	& \bfseries 0.1428	& 0.0612	&0.0428\\
and	& 0.0318	& 0.0641	& 0.0059
& at	& 0.1834	& 0.1881	& 0.1773\\
attached to	& 0.0275	& 0.0843	& 0.0556
& behind	& 0.1713	& 0.1884	& 0.1742\\
\bfseries belonging to	& \bfseries 0.4309	& 0.3044	& 0.0602
& between	& 0.0000	& 0.0000	& 0.0000\\
\bfseries carrying	& \bfseries 0.3638	& 0.2605	& 0.2565
& \bfseries covered in	& \bfseries 0.0595	& 0.0202	& 0.0274\\
covering	& 0.0650	& 0.0860	& 0.0970
& \bfseries eating	& \bfseries 0.4510 	& 0.1378	& 0.1253\\
flying in	& 0.0000	& 0.0000	& 0.0000
& for	& 0.0717	& 0.1079	& 0.0546\\
from	& 0.0000	& 0.0000	& 0.0000
& growing on	& 0.0000	& 0.0000	& 0.0000\\
hanging from	& 0.0519	& 0.0687	& 0.0418
& \bfseries has	& \bfseries 0.3008	& 0.2774	& 0.2874\\
\bfseries holding	& \bfseries 0.4510 	& 0.2579	& 0.1916
& \bfseries in	& \bfseries 0.1398	& 0.0994	& 0.0907\\
in front of	& 0.1751	& 0.2022	& 0.1713
& laying on	& 0.2496	& 0.2548	& 0.1681\\
looking at	& 0.2330 	& 0.2593	& 0.1502
& lying on	& 0.1240 	& 0.1378	& 0.0000\\
made of	& 0.0000	& 0.0000	& 0.0000
& mounted on	& 0.0841	& 0.1469	& 0.0000\\
near	& 0.2352	& 0.3536	& 0.1598
& of	& 0.2647	& 0.2781	& 0.2787\\
\bfseries on	& \bfseries 0.3079	& 0.1448	& 0.0640
& on back of	& 0.0235	& 0.0355	& 0.0142\\
over	& 0.0428	& 0.0761	& 0.0372
& \bfseries painted on	& \bfseries 0.1860 	& 0.0000	& 0.0000\\
\bfseries parked on	& \bfseries 0.2616	& 0.1872	& 0.1557
& \bfseries part of	& \bfseries 0.1595	& 0.1070	& 0.0000\\
\bfseries playing	& \bfseries 0.4165 	& 0.0000	& 0.0000
& \bfseries riding	& \bfseries 0.6510 	& 0.4367	& 0.4537\\
\bfseries says	& \bfseries 0.1332 	& 0.0000	& 0.0000
& \bfseries sitting on	& \bfseries 0.3475	& 0.2091	& 0.1769\\
standing on & 	0.2572	& 0.1842	& 0.0690
& to	& 0.0187	& 0.0246	& 0.0000\\
under	& 0.2218	& 0.2791	& 0.1419
& \bfseries using	& \bfseries 0.3281	& 0.2437	& 0.1690\\
\bfseries walking in	& \bfseries 0.0466	& 0.0141	& 0.0000
&\bfseries walking on	& \bfseries 0.1950	& 0.1084	& 0.0988\\
\bfseries watching	& \bfseries 0.4168	& 0.2326	& 0.2055
& \bfseries wearing	& \bfseries 0.5735	& 0.4443	& 0.4763\\
\bfseries wears	& \bfseries 0.4312	& 0.3787	& 0.0000
& \bfseries with	& \bfseries 0.1516	& 0.1289	& 0.0345\\

\hline
\end{tabular}}
\end{table*}

In Tab. \ref{tabx}, We show the \textbf{mR@100} metric performance on each relation class. The results fully demonstrate the superiority of LBL (performance on most fine-grained relations is better), which is significantly beneficial for subsequent reasoning tasks. The total results are \textbf{19.68\%}, 14.60\%, and 9.59\% for (TDE+OEM+LBL), (TDE+OEM), and (TDE), respectively.

\section{Discussion}
In this paper, based on scene graph generation tasks, we delve into the potential of our proposed logical bias learning (LBL) strategy for object relation prediction. Meanwhile, an effective object feature enhancement module is proposed. Through extensive experiments, we demonstrate the superiority of our methods. \\
Moreover, we firmly believe that these contributions will significantly bridge the gap between the SGG and practical applications, which can further be beneficial for two aspects: \textbf{1)} making it possible to automatically and efficiently generate high-quality cross-modal graph structural data, which can be used to pre-train foundation models; \textbf{2)} directly being involved in the process of object relation prediction of any model. However, our methods have only been evaluated on SGG tasks. Therefore, more experiments on these aspects are needed in the future.

\bibliography{iclr2023_conference}

\begin{thebibliography}{31}
\providecommand{\natexlab}[1]{#1}
\providecommand{\url}[1]{\texttt{#1}}
\expandafter\ifx\csname urlstyle\endcsname\relax
  \providecommand{\doi}[1]{doi: #1}\else
  \providecommand{\doi}{doi: \begingroup \urlstyle{rm}\Url}\fi

\bibitem[Ba et~al.(2016)Ba, Kiros, and Hinton]{ba2016layer}
Jimmy~Lei Ba, Jamie~Ryan Kiros, and Geoffrey~E Hinton.
\newblock Layer normalization.
\newblock \emph{arXiv preprint arXiv:1607.06450}, 2016.

\bibitem[Chen et~al.(2019)Chen, Yu, Chen, and Lin]{chen2019knowledge}
Tianshui Chen, Weihao Yu, Riquan Chen, and Liang Lin.
\newblock Knowledge-embedded routing network for scene graph generation.
\newblock In \emph{Proceedings of the CVPR}, 2019.

\bibitem[Dai et~al.(2017)Dai, Qi, Xiong, Li, Zhang, Hu, and Wei]{dai2017deformable}
Jifeng Dai, Haozhi Qi, Yuwen Xiong, Yi~Li, Guodong Zhang, Han Hu, and Yichen Wei.
\newblock Deformable convolutional networks.
\newblock In \emph{Proceedings of the ICCV}, 2017.

\bibitem[Dong et~al.(2022)Dong, Gan, Song, Wu, Cheng, and Nie]{dong2022stacked}
Xingning Dong, Tian Gan, Xuemeng Song, Jianlong Wu, Yuan Cheng, and Liqiang Nie.
\newblock Stacked hybrid-attention and group collaborative learning for unbiased scene graph generation.
\newblock In \emph{Proceedings of the CVPR}, 2022.

\bibitem[Geirhos et~al.(2018)Geirhos, Rubisch, Michaelis, Bethge, Wichmann, and Brendel]{geirhos2018imagenet}
Robert Geirhos, Patricia Rubisch, Claudio Michaelis, Matthias Bethge, Felix~A Wichmann, and Wieland Brendel.
\newblock Imagenet-trained cnns are biased towards texture; increasing shape bias improves accuracy and robustness.
\newblock \emph{arXiv preprint arXiv:1811.12231}, 2018.

\bibitem[Glymour et~al.(2016)Glymour, Pearl, and Jewell]{glymour2016causal}
Madelyn Glymour, Judea Pearl, and Nicholas~P Jewell.
\newblock \emph{Causal inference in statistics: A primer}.
\newblock John Wiley \& Sons, 2016.

\bibitem[Johnson et~al.(2015)Johnson, Krishna, Stark, Li, Shamma, Bernstein, and Fei-Fei]{johnson2015image}
Justin Johnson, Ranjay Krishna, Michael Stark, Li-Jia Li, David Shamma, Michael Bernstein, and Li~Fei-Fei.
\newblock Image retrieval using scene graphs.
\newblock In \emph{Proceedings of the CVPR}, 2015.

\bibitem[Li et~al.(2021)Li, Zhang, Wan, and He]{li2021bipartite}
Rongjie Li, Songyang Zhang, Bo~Wan, and Xuming He.
\newblock Bipartite graph network with adaptive message passing for unbiased scene graph generation.
\newblock In \emph{Proceedings of the CVPR}, 2021.

\bibitem[Li et~al.(2022)Li, Zhang, Bai, Zhao, Jiang, and Yuan]{li2022ppdl}
Wei Li, Haiwei Zhang, Qijie Bai, Guoqing Zhao, Ning Jiang, and Xiaojie Yuan.
\newblock Ppdl: Predicate probability distribution based loss for unbiased scene graph generation.
\newblock In \emph{Proceedings of the CVPR}, 2022.

\bibitem[Liang et~al.(2021)Liang, Jiang, and Liu]{liang2021graphvqa}
Weixin Liang, Yanhao Jiang, and Zixuan Liu.
\newblock Graphvqa: Language-guided graph neural networks for scene graph question answering.
\newblock \emph{NAACL-HLT 2021}, 2021.

\bibitem[Lin et~al.(2022)Lin, Ding, Zhang, Zhan, and Tao]{lin2022ru}
Xin Lin, Changxing Ding, Jing Zhang, Yibing Zhan, and Dacheng Tao.
\newblock Ru-net: regularized unrolling network for scene graph generation.
\newblock In \emph{Proceedings of the CVPR}, 2022.

\bibitem[Nguyen et~al.(2021)Nguyen, Tripathi, Du, Guha, and Nguyen]{nguyen2021defense}
Kien Nguyen, Subarna Tripathi, Bang Du, Tanaya Guha, and Truong~Q Nguyen.
\newblock In defense of scene graphs for image captioning.
\newblock In \emph{Proceedings of the ICCV}, 2021.

\bibitem[Pan et~al.(2022)Pan, Ge, Lu, Song, Chen, Huang, and Huang]{pan2022integration}
Xuran Pan, Chunjiang Ge, Rui Lu, Shiji Song, Guanfu Chen, Zeyi Huang, and Gao Huang.
\newblock On the integration of self-attention and convolution.
\newblock In \emph{Proceedings of the CVPR}, 2022.

\bibitem[Pearl \& Mackenzie(2018)Pearl and Mackenzie]{pearl2018book}
Judea Pearl and Dana Mackenzie.
\newblock \emph{The book of why: the new science of cause and effect}.
\newblock Basic books, 2018.

\bibitem[Ren et~al.(2015)Ren, He, Girshick, and Sun]{ren2015faster}
Shaoqing Ren, Kaiming He, Ross Girshick, and Jian Sun.
\newblock Faster r-cnn: Towards real-time object detection with region proposal networks.
\newblock \emph{Advances in neural information processing systems}, 28, 2015.

\bibitem[Simon(1990)]{simon1990bounded}
Herbert~A Simon.
\newblock Bounded rationality.
\newblock \emph{Utility and probability}, pp.\  15--18, 1990.

\bibitem[Tang et~al.(2019)Tang, Zhang, Wu, Luo, and Liu]{tang2019learning}
Kaihua Tang, Hanwang Zhang, Baoyuan Wu, Wenhan Luo, and Wei Liu.
\newblock Learning to compose dynamic tree structures for visual contexts.
\newblock In \emph{Proceedings of the CVPR}, 2019.

\bibitem[Tang et~al.(2020)Tang, Niu, Huang, Shi, and Zhang]{tang2020unbiased}
Kaihua Tang, Yulei Niu, Jianqiang Huang, Jiaxin Shi, and Hanwang Zhang.
\newblock Unbiased scene graph generation from biased training.
\newblock In \emph{Proceedings of the CVPR}, 2020.

\bibitem[Vaswani et~al.(2017)Vaswani, Shazeer, Parmar, Uszkoreit, Jones, Gomez, Kaiser, and Polosukhin]{vaswani2017attention}
Ashish Vaswani, Noam Shazeer, Niki Parmar, Jakob Uszkoreit, Llion Jones, Aidan~N Gomez, {\L}ukasz Kaiser, and Illia Polosukhin.
\newblock Attention is all you need.
\newblock \emph{Advances in neural information processing systems}, 2017.

\bibitem[Xie et~al.(2017)Xie, Girshick, Doll{\'a}r, Tu, and He]{xie2017aggregated}
Saining Xie, Ross Girshick, Piotr Doll{\'a}r, Zhuowen Tu, and Kaiming He.
\newblock Aggregated residual transformations for deep neural networks.
\newblock In \emph{Proceedings of the CVPR}, 2017.

\bibitem[Xu et~al.(2017)Xu, Zhu, Choy, and Fei-Fei]{xu2017scene}
Danfei Xu, Yuke Zhu, Christopher~B Choy, and Li~Fei-Fei.
\newblock Scene graph generation by iterative message passing.
\newblock In \emph{Proceedings of the CVPR}, 2017.

\bibitem[Xu et~al.(2021)Xu, Zhang, Zhang, and Tao]{xu2021vitae}
Yufei Xu, Qiming Zhang, Jing Zhang, and Dacheng Tao.
\newblock Vitae: Vision transformer advanced by exploring intrinsic inductive bias.
\newblock \emph{Advances in Neural Information Processing Systems}, 34:\penalty0 28522--28535, 2021.

\bibitem[Yan et~al.(2020)Yan, Shen, Jin, Huang, Jiang, Chen, and Hua]{yan2020pcpl}
Shaotian Yan, Chen Shen, Zhongming Jin, Jianqiang Huang, Rongxin Jiang, Yaowu Chen, and Xian-Sheng Hua.
\newblock Pcpl: Predicate-correlation perception learning for unbiased scene graph generation.
\newblock In \emph{Proceedings of the 28th ACM International Conference on Multimedia}, 2020.

\bibitem[Yang et~al.(2022)Yang, Ang, Guo, Zhou, Zhang, and Liu]{yang2022panoptic}
Jingkang Yang, Yi~Zhe Ang, Zujin Guo, Kaiyang Zhou, Wayne Zhang, and Ziwei Liu.
\newblock Panoptic scene graph generation.
\newblock In \emph{Proceedings of the ECCV}. Springer, 2022.

\bibitem[Yang et~al.(2019)Yang, Tang, Zhang, and Cai]{yang2019auto}
Xu~Yang, Kaihua Tang, Hanwang Zhang, and Jianfei Cai.
\newblock Auto-encoding scene graphs for image captioning.
\newblock In \emph{Proceedings of the CVPR}, 2019.

\bibitem[Yin et~al.(2018)Yin, Sheng, Liu, Yu, Wang, Shao, and Loy]{yin2018zoom}
Guojun Yin, Lu~Sheng, Bin Liu, Nenghai Yu, Xiaogang Wang, Jing Shao, and Chen~Change Loy.
\newblock Zoom-net: Mining deep feature interactions for visual relationship recognition.
\newblock In \emph{Proceedings of the ECCV}, 2018.

\bibitem[Zareian et~al.(2020)Zareian, Karaman, and Chang]{zareian2020bridging}
Alireza Zareian, Svebor Karaman, and Shih-Fu Chang.
\newblock Bridging knowledge graphs to generate scene graphs.
\newblock In \emph{Proceedings of the ECCV}. Springer, 2020.

\bibitem[Zellers et~al.(2018)Zellers, Yatskar, Thomson, and Choi]{zellers2018neural}
Rowan Zellers, Mark Yatskar, Sam Thomson, and Yejin Choi.
\newblock Neural motifs: Scene graph parsing with global context.
\newblock In \emph{Proceedings of the CVPR}, 2018.

\bibitem[Zhang et~al.(2022)Zhang, Yao, Chen, Ji, Liu, Sun, and Chua]{zhang2022fine}
Ao~Zhang, Yuan Yao, Qianyu Chen, Wei Ji, Zhiyuan Liu, Maosong Sun, and Tat-Seng Chua.
\newblock Fine-grained scene graph generation with data transfer.
\newblock In \emph{Proceedings of the ECCV}. Springer, 2022.

\bibitem[Zhou et~al.(2022)Zhou, Li, Chen, and Zhu]{zhou2022disentangled}
Xinyu Zhou, Shilin Li, Huen Chen, and Anna Zhu.
\newblock Disentangled ocr: A more granular information for “text”-to-image retrieval.
\newblock In \emph{Pattern Recognition and Computer Vision: 5th Chinese Conference, PRCV 2022, Shenzhen, China, November 4--7, 2022, Proceedings, Part I}. Springer, 2022.

\bibitem[Zhou et~al.(2023)Zhou, Zhu, Chen, and Pan]{zhou2023scene}
Xinyu Zhou, Anna Zhu, Huen Chen, and Wei Pan.
\newblock Scene text involved" text"-to-image retrieval through logically hierarchical matching.
\newblock In \emph{2023 IEEE International Conference on Multimedia and Expo}. IEEE, 2023.

\end{thebibliography}
\bibliographystyle{iclr2023_conference}

\end{document}